%% file: main.tex
\documentclass{article}

\usepackage[preprint]{neurips_2026}

\usepackage[utf8]{inputenc}
\usepackage[T1]{fontenc}
\usepackage{amsmath,amssymb,amsfonts}
\usepackage{tikz-cd}
\usepackage{multicol}
\usepackage{booktabs}
\usepackage{nicefrac}
\usepackage{microtype}
\usepackage{hyperref}
\hypersetup{hidelinks}
\usepackage{url}
\setcitestyle{authoryear,round}

\usepackage[none]{hyphenat}
\usepackage{float}
\title{CohortFlow: A Steerable Clinical Cohort Curation Agent with Localized Semantic Correction}

\author{
  Neil Chen\thanks{\texttt{neilchen@berkeley.edu}} \\
  UC Berkeley
  \And
  Adrian Serapio \\
  UC Berkeley; UC San Francisco
  \And
  Hanxue Gu \\
  UC San Francisco
  \AND
  Evan Lee \\
  UC Berkeley
  \And
  Kang Wang \\
  UC San Francisco
  \And
  Yang Yang \\
  UC San Francisco
}

\begin{document}
\maketitle

\begin{abstract}
Language model agents for clinical cohort curation should accommodate evolving eligibility criteria without repeatedly processing unchanged requirements. We introduce \href{https://github.com/NEILCHENHENRI/cohortflow}{CohortFlow}, a steerable agent that represents natural language corrections as localized updates to a structured cohort specification and reconciles them with a persistent execution graph. Explicitly recorded semantic dependencies enable affected computations to be selectively invalidated while preserving valid prior results. On 100 scenarios derived from clinical-trial eligibility criteria, comprising 126 correction turns, localized steering achieves 77.0\% requirement micro-F1 versus 76.4\% for full regeneration using the same 8B model, a small numerical advantage that is not statistically significant ($p=0.493$). It reduces mean correction latency by 64.8\% after completion and 44.3\% during execution, with 38.1\% fewer model calls. In a separate proof-of-concept evaluation over MIMIC, all 13 cohort states execute successfully, while correction trajectories demonstrate selective preservation of unaffected computations and recomputation of changed or dependent nodes. Together, these results show that localized semantic steering can substantially reduce correction latency and model calls while maintaining strong requirement-level accuracy.
\end{abstract}

\input{sections/introduction}

\input{sections/related_work}

\input{sections/methods}

\input{sections/experiments}

\input{sections/results}

\input{sections/disc}

\input{sections/conclusion}

\newpage
\bibliographystyle{plainnat}
\bibliography{references}

\newpage
\input{sections/appendix}

\end{document}

%% file: sections/introduction.tex
\section{Introduction}

Clinical cohort curation is often iterative: an initial set of eligibility criteria may be revised as clinicians inspect intermediate results, clarify intent, or refine the target population. Existing systems increasingly use language models to translate natural language criteria into computable cohort definitions or reason over electronic health records (EHRs) \citep{yuan2019criteria2query,park2024criteria2query3,shi2024ehragent}. However, conventional pipelines generally treat a revised request as a new task, potentially repeating interpretation and computation even when most of the existing cohort definition remains unchanged.

This limitation becomes particularly important when corrections arrive during execution. A clinician may add an exclusion criterion, modify a laboratory threshold, or revise a temporal relationship after other parts of the cohort have already been processed. Restarting from the revised request discards valid intermediate work, while modifying only the final result risks leaving the execution state inconsistent with the revised intent. Importantly, execution order does not determine semantic dependency: changing an earlier computation need not invalidate later computations that do not depend on it.

We introduce a steerable cohort-curation agent that incorporates natural language corrections without restarting the entire task. The agent maintains an explicit cohort specification alongside execution state, allowing revisions to propagate according to semantic dependencies rather than execution order. This enables corrections to be incorporated during execution while preserving computation that remains valid.

Our contributions are:
\begin{itemize}
    \item We formulate \textbf{semantic mid-run steering for clinical cohort curation} and introduce an incremental \textbf{Task Delta} mechanism that interprets natural language corrections as localized changes to an existing cohort specification while preserving unchanged structured state.

    \item We introduce \textbf{semantic validity-aware execution across cohort revisions}, using stable requirement references and semantic fingerprints to selectively invalidate affected computation while retaining valid artifacts across structured database operations and unstructured clinical-note reasoning.

    \item We develop a \textbf{stateful steering framework for heterogeneous EHR workflows} that supports resumable execution, ordered interventions, and sequential or parallel correction scheduling, and evaluate its semantic correctness, computation reuse, and adaptation efficiency.
\end{itemize}

%% file: sections/related_work.tex
\section{Related Work}

\subsection{Interactive and Steerable LLM Agents}

Recent systems have moved beyond transactional agent interaction by allowing users to intervene during execution. CowPilot supports interleaved human and agent actions during web navigation, while InterDeepResearch exposes intermediate research processes for inspection and steering \citep{huq2025cowpilot,pan2026interdeepresearch}. Plover exposes persistent plans that users can inspect and revise, allowing localized corrections to influence subsequent execution without discarding prior progress \citep{venkatesan2026plover}. These systems demonstrate the value of exposing persistent intermediate state for user intervention.

MUSE is particularly related to our work in reconstructing low-level data-science agent traces into semantic workflow representations through which users can reference and revise specific steps \citep{chen2026muse}. Revisable by Design considers user interventions during streaming agent execution, while BoardroomAI compiles human interventions into structured updates to an evolving decision graph \citep{zhai2026revisable,manivannan2026boardroomai}. Our work instead treats the user's evolving cohort definition itself as persistent computational state. Corrections are represented as localized Task Deltas over this specification, such that subsequent execution is defined with respect to the revised clinical intent rather than an independently edited plan, workflow, or decision graph.

\subsection{Dependency-Aware Execution and Reuse}

Beyond representing user revisions, a complementary line of work studies how existing computation should be treated after a change. Revisable by Design uses Earliest-Conflict Rollback to preserve computation preceding the earliest conflicting action, while BoardroomAI propagates interventions through dependencies in its decision graph to preserve unaffected artifacts and selectively reactivate affected computation \citep{zhai2026revisable,manivannan2026boardroomai}. Execution Lineage represents agentic computation as a graph of artifact-producing operations with explicit dependencies, Shepherd provides reversible execution traces supporting replay, revert, and fork, and AgentTether uses dependency-aware trajectory representations to localize failures and selectively recover affected execution regions \citep{rosen2026executionlineage,yu2026shepherd,zhao2026agenttether}.

Our setting poses a different validity problem: when the semantics of the task itself change, existing computation must be reconsidered with respect to the revised specification rather than solely its position in an execution trace or graph. In cohort curation, this validity may occur at different granularities across structured database results, retrieved evidence, and patient-level note judgments.

\subsection{LLMs for EHR Cohort Curation and Phenotyping}

Clinical cohort construction has long motivated natural language interfaces to structured health data. Criteria2Query translates free-text eligibility criteria into structured cohort definitions and executable database queries \citep{yuan2019criteria2query}. Criteria2Query 2.0 introduces human interaction for correcting criteria selection, parsing, and concept mapping, while Criteria2Query 3.0 incorporates generative LLMs for concept extraction, query generation, and reasoning \citep{fang2022criteria2query,park2024criteria2query3}. More recent systems extend LLM-based cohort construction and eligibility reasoning: EC2Seq2Sql maps eligibility criteria through structured representations to executable EHR queries, GDC Cohort Copilot supports interactive cohort refinement over cancer genomic data, and TrialGPT retrieves patient evidence for criterion-level eligibility assessment \citep{liu2026ec2seq2sql,song2025gdccohortcopilot,jin2024trialgpt}.

A complementary line of work reasons directly over heterogeneous patient data. EHRAgent combines LLM reasoning with iterative code execution over EHR tables, while targeted LLM-based phenotyping combines structured signals with retrieval and language-model reasoning over clinical text, limiting expensive note analysis to patients insufficiently resolved by structured evidence \citep{shi2024ehragent,owens2026targeted}. These systems establish many components required for LLM-assisted cohort curation, including structured cohort representations, interactive refinement, database execution, and reasoning over unstructured clinical evidence.

Our work instead focuses on maintaining consistency between an evolving cohort definition and ongoing computation when that definition is revised during execution. Rather than treating a revised cohort as a new query, we treat the evolving specification and its existing execution state as a persistent computation whose validity must be maintained as clinical intent changes.

%% file: sections/methods.tex
\section{Methods}

\subsection{System Overview}

We formulate cohort curation as a stateful computation over a structured cohort specification. A natural language request is parsed into a database-independent \texttt{TaskSpec} containing typed eligibility requirements, Boolean aggregation logic, and temporal relations. Clinical predicates are grounded to identifiers in MIMIC-IV dataset \citep{johnson2023mimiciv} and compiled into deterministic database operations, while free-text criteria are evaluated through a retrieval and adjudication pipeline.

The \texttt{TaskSpec} is compiled into an execution directed acyclic graph (DAG), whose intermediate artifacts persist across interactions. User corrections are compiled incrementally into the current specification: only newly introduced or modified requirements are parsed and grounded, while unchanged requirements retain their existing representations. The execution graph then determines which previous computations remain valid under the revised specification. The same state model supports sequential steering and parallel correction interpretation.

\begin{figure}[htbp]
    \centering
    \includegraphics[width=\linewidth]{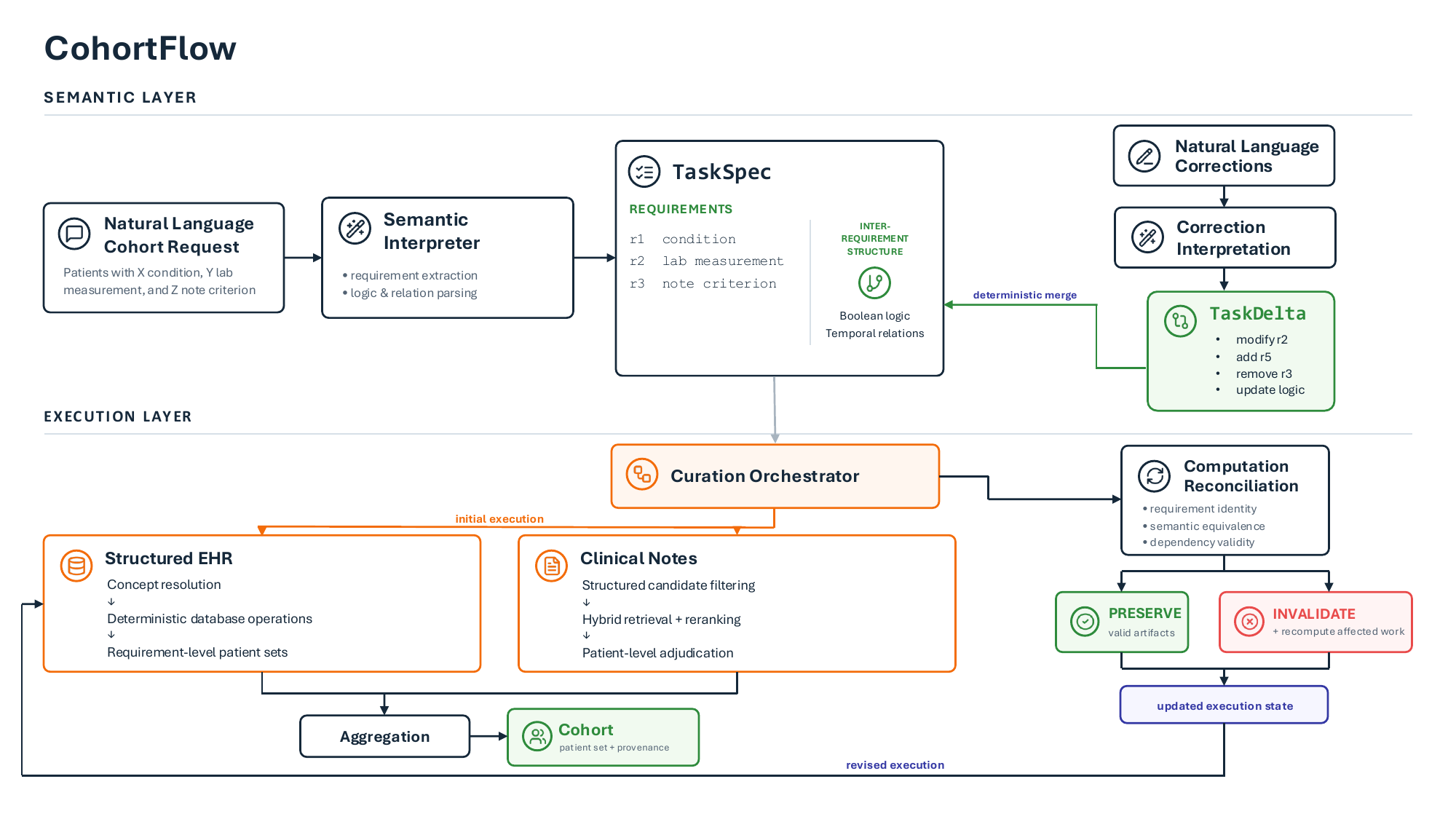}
    \caption{Architecture of the steerable clinical cohort curation system, showing initial cohort specification and execution, localized correction, and selective reconciliation of affected computations.}
    \label{fig:system}
\end{figure}

\subsection{Clinical Concept Resolution and Structured Execution}

Clinical expressions are grounded directly to MIMIC-IV source vocabularies. Conditions and procedures are mapped to ICD dictionary entries, laboratory measurements to MIMIC laboratory-item identifiers, and medications to drug names represented in prescriptions.

Resolution first attempts a unique normalized exact match. Otherwise,
candidates are retrieved using an inverted token index and ranked by an
IDF-weighted overlap score,

\[
S(q,d)=
\sum_{w\in\mathcal{T}(q)\cap\mathcal{T}(d)}
\log\left(1+\frac{N}{\mathrm{df}(w)}\right)
-\epsilon|\mathcal{T}(d)|,
\]

where $q$ is the clinical query, $d$ is a candidate dictionary entry,
$\mathcal{T}(\cdot)$ denotes the set of normalized non-stopword tokens,
$N$ is the number of entries in the corresponding MIMIC dictionary,
$\mathrm{df}(w)$ is the number of entries containing token $w$, and
$\epsilon=10^{-3}$ is a small length-penalty coefficient. The IDF-style
weighting assigns greater importance to informative tokens that occur in
fewer dictionary entries, while the small length penalty weakly favors
concise candidate descriptions.

If lexical retrieval fails, a medical LLM (e.g., MedGemma) generates canonical and equivalent clinical expressions that are used to repeat deterministic dictionary lookup. When a clinical concept maps to multiple valid codes, a patient satisfying any of the selected codes is considered to satisfy the requirement.

Instead of using an LLM, grounded requirements are compiled to parameterized SQL by deterministic operators. The execution layer supports quantitative constraints, event counts, admission attributes, and temporal \textsc{before}/\textsc{after} relations between clinical events.

\subsection{Clinical Note Retrieval and Adjudication}

Free-text criteria are evaluated using a hybrid retrieval pipeline. When permitted by the Boolean structure, structured predicates are evaluated first to restrict note retrieval to patients for whom the note criterion can still affect cohort membership.

Clinical notes are chunked by section and indexed for hybrid retrieval. Sparse candidates are retrieved using BM25, while dense candidates are retrieved using normalized semantic embeddings (e.g., BGE). Their union is subsequently scored using a cross-encoder reranker. For each candidate passage $i$, the reranker produces a logit $z_i$, which is converted to a relevance score
\[
s_i = \sigma(z_i),
\]
where $\sigma(\cdot)$ denotes the sigmoid function.

Evidence is filtered using a configurable fixed relevance threshold $\tau$, retaining passages with $s_i \geq \tau$. Candidate limits, if any, are applied separately within each patient, without a population-relative cutoff or shared retention cap. Consequently, changing the candidate population does not change the evidence retained for an existing patient, provided the query, source data, models, and retrieval settings remain unchanged. Retained passages are grouped by patient and passed to an LLM for final criterion adjudication.

\subsection{Mid-Execution Steering}

Corrections are applied as localized transformations of the current \texttt{TaskSpec}. Given state $T_t$ and correction $c_t$, an LLM produces a semantic delta $\Delta_t$ describing the requested additions, removals, or modifications. Only newly introduced or modified requirements are compiled through the same typed parsing and grounding mechanisms used for the initial request. Unchanged requirements are not regenerated. The compiled delta is then deterministically merged into the existing specification:

\[
(T_t,c_t)
\xrightarrow{\mathrm{interpret}}
\Delta_t
\xrightarrow{\mathrm{compile\ changed\ components}}
\widehat{\Delta}_t
\xrightarrow{\mathrm{merge\ with}\ T_t}
T_{t+1}.
\]

Unmodified requirements retain their structured representations and stable identifiers. Thus, correction is incremental at the specification level before dependency-aware reuse is considered at the execution level. Corrections are processed in semantic FIFO order: a subsequent correction is interpreted only after the preceding update has materialized its new \texttt{TaskSpec}. This prevents corrections from being applied against stale state.

Long-running deterministic operations expose safe interruption boundaries and preserve completed work, allowing compatible computation to resume after a correction.

\subsection{Sequential and Parallel Steering}

The system supports two scheduling policies over the same ordered state transitions. In \textbf{sequential mode}, a correction waits for the active semantic action before interpretation and application.

In \textbf{parallel mode}, the first queued correction may be interpreted concurrently with an active LLM action or supported human interaction. The resulting update is not immediately committed: it waits for the active action to complete and is then applied to the appropriate materialized state. Subsequent corrections remain queued, preserving semantic ordering.

Parallel steering therefore overlaps correction-interpretation latency without parallelizing the execution DAG or changing the committed sequence of \texttt{TaskSpec} states.

\subsection{Dependency-Aware Invalidation and Reuse}

Each execution node is assigned a semantic fingerprint derived from its operation, dependencies, and execution configuration. After a correction, the updated DAG is reconciled with the previous graph. A completed node $v$ is reusable when its stable identity and semantic fingerprint are unchanged, its required artifacts remain available, and all of its parent computations remain reusable. Conceptually,

\[
\operatorname{reuse}(v)
=
\operatorname{same}(v)
\land
\operatorname{available}(v)
\land
\bigwedge_{p\in\operatorname{parents}(v)}
\operatorname{reuse}(p),
\]

where $\operatorname{same}(v)$ denotes preservation of the node's semantic identity and fingerprint. Consequently, a semantic change invalidates only computations whose own semantics or upstream dependencies have changed rather than all computation occurring after the revision point.

Note processing additionally uses stage-specific reuse. Corpus-level chunks, indexes, and embeddings persist independently of individual note requirements. Patient-level adjudication is keyed by the canonical note requirement and retrieval configuration. If an unchanged note criterion is applied to a smaller candidate population, existing decisions for surviving patients are reused; if the population expands, only new patients are processed. Changes to the note semantics or retrieval policy invalidate the corresponding requirement-specific results.

This combination of graph-level reconciliation and patient-level note reuse allows steering to recompute only state affected by a semantic change.

%% file: sections/experiments.tex
\section{Experiments}

\subsection{Experimental Setup}

\paragraph{Benchmark construction.} We construct 100 cohort-curation scenarios using the training set of CohortX Task 1 \citep{alzahrani2026cohortx} as source material. Each scenario is derived from a distinct study and contains an initial clinician-style cohort request followed by zero or more corrections. Studies and criteria were selected heuristically to provide variation in clinical content, task complexity, and opportunities for meaningful cohort revision. Scenarios were manually curated with assistance from Claude Code and annotated using the same \texttt{TaskSpec} schema consumed by our execution framework, which represents cohort requirements as typed predicates together with their Boolean composition and optional temporal relations. Each scenario provides a gold TaskSpec for the initial request and for each subsequent correction. The benchmark contains 126 correction turns, including 26 multi-correction scenarios and eight intentionally ambiguous corrections requiring clarification. Corrections span additions, modifications, removals, replacements, and intentionally ambiguous requests requiring clarification. All scenarios were manually reviewed and approved against their source eligibility criteria, with schema and transition consistency additionally validated automatically.

\paragraph{Evaluation design.} We divide evaluation into two stages. Stage 1 evaluates semantic interpretation and steering against the gold \texttt{TaskSpec}s, while Stage 2 evaluates downstream cohort execution and correction-aware reuse from frozen structured specifications. Stage 1 systems use Qwen3-8B with identical inference settings: \texttt{bitsandbytes} 4-bit NF4 quantization with double quantization, \texttt{bfloat16} computation, temperature 0, a maximum generation length of 1,536 tokens, and reasoning disabled. The evaluation runs on an NVIDIA L4 GPU using \texttt{transformers} 4.57.6 and \texttt{bitsandbytes} 0.50.2.

\subsection{Stage 1: Semantic Steering}

We compare our steerable agent against a naive restart baseline modeled after the full-generation paradigm used in existing natural language cohort-query systems \citep{park2024criteria2query3}. For the initial request, both systems construct a complete \texttt{TaskSpec}. Upon receiving a correction, the steerable agent interprets the requested semantic change and reconciles it with the existing \texttt{TaskSpec}, whereas the naive baseline restarts semantic processing and regenerates the complete specification from the updated request history. We disable full-regeneration fallback for the steerable agent so that its performance reflects localized steering rather than hidden restarts. Both systems receive identical requests and use the same model and inference settings.

We evaluate corrections under both post-completion and mid-run conditions. Post-completion corrections are submitted after the current specification has been produced. For mid-run steering, the benchmark harness injects the correction during predicate compilation while execution is still active. Across the benchmark, 111 correction turns are evaluated mid-run and 15 after completion. For multi-correction scenarios, corrections are processed in order so that each resulting \texttt{TaskSpec} becomes the state against which the next correction is applied.

Semantic accuracy is evaluated against the gold \texttt{TaskSpec} after each turn. We report micro-averaged requirement precision, recall, and F1, together with Boolean logic accuracy, conditional temporal relation accuracy, structural validity, and clarification accuracy for intentionally ambiguous corrections. Predicted and gold requirements are aligned one-to-one within each requirement type to maximize shared structured facts, with lexical similarity used only to break ties. Micro-F1 pools exact matches of kind, polarity, required status, and individual payload fields after text normalization (Unicode, case, punctuation, and whitespace), with no synonym expansion for simplicity or numerical tolerance and with comparison operators preserved exactly. Relation accuracy is evaluated only when the predicates required to express the gold relation are present and alignable, preventing an upstream requirement error from being counted again as a relation error. Efficiency is measured using wall-clock latency, number of LLM calls, and prompt, completion, and total token counts.

\subsection{Stage 2: Backend Execution}

Stage 2 evaluates whether the structured representations produced by the semantic layer are executable by the downstream cohort-curation framework. To isolate backend behavior from Stage 1 parsing errors, we use manually frozen valid \texttt{TaskSpec}s rather than model-generated specifications. Ten proof-of-concept scenarios cover condition, patient, procedure, laboratory, medication, admission, Boolean, and temporal operations and are executed through concept resolution and database evaluation over MIMIC. Concept disambiguation is automated by selecting the highest-ranked candidate when confirmation is required.

Three scenarios additionally receive pre-specified structured corrections, yielding 13 evaluated cohort states in total. These corrections respectively add a new condition, modify a laboratory threshold, and modify a temporal constraint, allowing us to examine distinct reconciliation behaviors. Corrections are applied as frozen \texttt{TaskDelta}s to the existing execution state rather than restarting execution. We record whether each state successfully materializes a cohort and inspect the resulting execution graph to determine whether prior computations are correctly reused or invalidated. We use this stage to evaluate backend executability and correction-aware graph reconciliation rather than clinical correctness of the resulting cohorts.

%% file: sections/results.tex
\section{Results}

\subsection{Semantic Accuracy}

Localized steering largely preserves requirement-level semantic accuracy relative to full regeneration. Across jointly scoreable correction states with gold \texttt{TaskSpec}s, the steerable agent achieves 77.0\% requirement micro-F1, compared with 76.4\% for the naive restart baseline, a difference of $+0.6$ percentage points. The corresponding precision and recall are 73.4\% and 81.0\% for the steerable agent and 73.0\% and 80.0\% for the restart baseline. The difference remains small across benchmark difficulty: steerable versus restart micro-F1 is 76.9\% versus 77.2\% on simple scenarios, 80.0\% versus 78.5\% on medium scenarios, and 74.9\% versus 74.8\% on hard scenarios.

\begin{table}[h]
\centering
\caption{Semantic accuracy on correction states. For each metric, $N$ denotes the number of correction states included in the comparison. Boolean logic and temporal relation accuracy are evaluated only on states scoreable for both systems.}
\label{tab:semantic_results}
\small
\setlength{\tabcolsep}{4pt}
\begin{tabular}{@{}lrrrr@{}}
\toprule
Metric & Steerable & Restart & $N$ & Paired $p$ \\
\midrule
Requirement precision (\%) & 73.4 & 73.0 & 110 & -- \\
Requirement recall (\%) & 81.0 & 80.0 & 110 & -- \\
Requirement micro-F1 (\%) & 77.0 & 76.4 & 110 & 0.493 \\
Boolean logical equivalence (\%) & 85.7 & 94.3 & 35 & 0.500 \\
Temporal relation accuracy (\%) & 90.5 & 89.5 & 95 & 1.000 \\
Valid outputs / planned corrections (\%) & 94.4 & 99.2 & 126 & -- \\
Ambiguity clarification accuracy (\%) & 12.5 & 50.0 & 8 & -- \\
\bottomrule
\end{tabular}
\end{table}

Because a scenario may contain multiple successive corrections, its correction states are not independent observations. We therefore estimate uncertainty by repeatedly resampling entire scenarios rather than individual correction turns. In each of 100,000 bootstrap samples, all jointly scoreable correction turns belonging to a selected scenario are retained, and the same sampled scenarios are used for both systems to preserve the paired comparison. Requirement true positives, false positives, and false negatives are pooled to recompute micro-F1 and the steerable-minus-restart difference. The resulting bootstrap distribution gives a 95\% confidence interval of $[-1.23, 2.45]$ percentage points for the observed $+0.6$ percentage-point difference. The observed difference favors the steerable agent, although the confidence interval includes zero and therefore does not establish a statistically significant advantage.

We additionally assess statistical significance using a paired case-cluster randomization test. Under the null hypothesis that the two systems are exchangeable, system labels are swapped within entire scenarios, preserving all correction turns from each scenario as a cluster. For requirement micro-F1, we use 100,000 randomizations and compute the two-sided $p$-value from the proportion of randomized absolute differences at least as large as the observed difference, with the standard $+1$ correction. Because the comparison is paired by scenario, these $p$-values reflect whether one system consistently differs from the other on the same scenarios rather than simply the magnitude of the aggregate percentage-point difference. We use $p<0.05$ as the threshold for statistical significance. For requirement micro-F1, $p=0.493$. Thus, the small observed advantage in requirement micro-F1 is not statistically significant.

Boolean composition and temporal relations are evaluated on jointly scoreable correction states. Boolean scoring requires complete predicate alignment and compares logical equivalence, while temporal scoring requires alignment of the referenced predicates and compares relation lists using aligned endpoints, ignoring generated relation identifiers. Structures affected by missing or corrupted upstream predicates are excluded. Table~\ref{tab:semantic_results} and the paired statistical comparisons use the same jointly scoreable states for each metric.

For temporal relations, accuracy is 90.5\% for the steerable agent and 89.5\% for restart across 95 jointly scoreable correction states. The paired bootstrap gives a 95\% confidence interval of $[-5.49, 8.16]$ percentage points, while the paired randomization test gives $p=1.000$. Both results provide insufficient evidence of a statistically significant difference.

Boolean composition remains more challenging for localized steering. Logical equivalence accuracy is 85.7\% for the steerable agent and 94.3\% for full regeneration across 35 jointly scoreable states, a $-8.6$ percentage-point difference. The paired bootstrap gives a 95\% confidence interval of $[-23.53, 3.13]$ percentage points, while the paired randomization test gives $p=0.500$. Neither rejects the null hypothesis at the 0.05 level. Given the small jointly scoreable sample (35 states across 32 scenarios), the observed gap should be interpreted cautiously.

Additionally, on the eight intentionally ambiguous corrections, the steerable agent requested the expected clarification in 1/8 cases, compared with 4/8 for the restart baseline, highlighting clarification handling as an additional limitation of the current localized correction path. Across all 126 planned correction states, valid-output coverage is 94.4\% for the steerable agent and 99.2\% for restart. This denominator includes the two unattempted steerable corrections as failures.

\subsection{Steering Efficiency}

Localized steering substantially reduces correction latency and the number of model invocations. For post-completion corrections, mean wall-clock latency decreases from 48.6\,s with full regeneration to 17.1\,s with steering, corresponding to a 64.8\% reduction and a $2.84\times$ speedup. For mid-run corrections, latency decreases from 69.7\,s to 38.8\,s, a 44.3\% reduction and a $1.80\times$ speedup. Median latencies show the same pattern, decreasing from 47.3\,s to 17.0\,s post-completion and from 67.6\,s to 41.3\,s mid-run.

\begin{table}[h]
\centering
\caption{Correction efficiency under steerable and restart execution.}
\label{tab:efficiency_results}
\begin{tabular}{lrr}
\toprule
Metric & Steerable & Restart \\
\midrule
Mean post-completion latency (s) & 17.1 & 48.6 \\
Median post-completion latency (s) & 17.0 & 47.3 \\
Mean mid-run latency (s) & 38.8 & 69.7 \\
Median mid-run latency (s) & 41.3 & 67.6 \\
LLM calls per correction & 1.90 & 3.07 \\
Completion tokens per correction & 222 & 680 \\
Total tokens per correction & 4,655 & 3,996 \\
\bottomrule
\end{tabular}
\end{table}

The steerable agent requires approximately 38.1\% fewer LLM calls per correction and 67.3\% fewer completion tokens. This reduction does not translate into lower total token usage: the steerable agent uses approximately 16.5\% more total tokens because localized correction requires a larger prompt context. Thus, the primary efficiency benefit of steering is reduced latency and model invocation rather than reduced total token consumption. Paired scenario-level bootstrap intervals place the mean latency reductions at 61.9--67.5\% after completion and 41.6--47.0\% during execution (95\% confidence). Together with the semantic results, these findings indicate that localized steering avoids substantial recomputation while achieving slightly higher observed requirement micro-F1.

\subsection{Backend Execution and Reuse}

All 13 Stage-2 cohort states successfully complete backend execution and materialize a patient cohort, demonstrating executability across the tested condition, patient, procedure, laboratory, medication, admission, Boolean, and temporal operations.

The three correction scenarios further demonstrate selective reconciliation of the execution graph. When a new COPD condition is added to an existing cirrhosis cohort, the previously computed cirrhosis evaluation is preserved while a new COPD branch is evaluated and downstream aggregation is recomputed. When the potassium threshold is changed from $>5.5$ to $>6.0$, the affected laboratory predicate and its downstream computations are invalidated rather than reused. Finally, when a temporal constraint between elevated potassium and paracentesis is widened from 7 to 14 days, both underlying predicate evaluations are preserved while the temporal relation and downstream cohort computations are recomputed.

These cases demonstrate that correction handling operates below the level of full workflow restart: computations unaffected by a semantic change can remain valid, while computations dependent on the changed specification are selectively invalidated. Because Stage 2 contains only three correction trajectories, these results should be interpreted as a proof of concept for correction-aware reuse rather than a comprehensive quantitative evaluation of reuse efficiency. We do not separately ablate execution reuse, since removing reuse directly reduces the system to full recomputation and provides limited additional insight beyond the reuse behavior demonstrated above.

%% file: sections/disc.tex
\section{Discussion and Limitations}

Our results suggest that semantic steering can reduce the cost of revising an executing cohort-curation workflow without requiring every correction to restart from the original request. Requirement-level accuracy is similar to full regeneration, while localized steering substantially reduces correction latency and model invocations. Stage 2 further illustrates the intended execution behavior: revisions can preserve computations whose semantic dependencies remain unchanged while selectively recomputing affected portions of the graph. However, the lower Boolean-logic accuracy and poor clarification performance also expose an important limitation of localized revision. Preserving individual requirements does not guarantee that their global composition remains correct, particularly when a correction changes how existing predicates interact.

Our evaluation has several limitations. Stage 1 uses a single 8B language model under 4-bit quantization. Quantization may constrain model performance, and stronger or higher-precision models may improve correction interpretation and potentially narrow the observed accuracy gap between localized steering and full regeneration. Evaluating this possibility across model families and scales remains future work. Although the benchmark is grounded in real clinical-trial eligibility criteria, its correction trajectories are constructed rather than collected prospectively from clinicians. In addition, Stage 1 excludes clinical-note retrieval and reasoning. Because these operations introduce substantially more retrieval and model computation than structured specification processing, we expect selective reuse to provide greater latency benefits when the full heterogeneous workflow is exercised, but this remains to be evaluated empirically.

Stage 2 is intentionally a proof of concept: its 13 cohort states demonstrate backend executability and representative reuse and invalidation behaviors, but do not constitute a comprehensive evaluation of reuse efficiency or clinical cohort validity. Future work should therefore evaluate the framework with additional models, larger and more diverse EHR datasets, full note-processing workloads, and corrections collected through clinician interaction.

Finally, successful execution should not be interpreted as evidence that a generated cohort is clinically correct. Errors in requirement interpretation, ambiguity handling, concept resolution, or clinical-note reasoning can propagate to cohort membership even when the resulting execution graph is structurally valid. The system is therefore intended to support, rather than replace, expert cohort definition and review.

%% file: sections/conclusion.tex
\section{Conclusion}

We presented a steerable agent for clinical cohort curation that incorporates natural language corrections through localized specification updates and semantic validity-aware execution. By maintaining structured revision state and reconciling changes with an explicit execution graph, the framework can preserve valid prior computation rather than restarting the entire workflow after each intervention. Our evaluation shows that localized steering maintains comparable requirement-level semantic accuracy while substantially reducing correction latency and model invocations, and our backend experiments demonstrate selective reuse across representative cohort revisions. These results motivate correction-aware execution as a practical direction for interactive clinical data agents.

%% file: sections/appendix.tex
\appendix

\section{Code Availability}
The implementation is available at
\url{https://github.com/NEILCHENHENRI/cohortflow}.

\section{Example of Mid-Run Cohort Revision}
\label{app:steering-example}

We provide an end-to-end example illustrating how a natural language correction
changes the cohort specification and how unaffected computation is preserved
during revision.

\subsection{Initial Request and Task Specification}

The initial request is:

\begin{quote}
\textit{``Find patients with type 2 diabetes or hypertension.''}
\end{quote}

The request is represented by the following \texttt{TaskSpec}. Stable predicate
and group references make the logical structure explicit: predicates
\texttt{p1} and \texttt{p2} are combined by the disjunction \texttt{g1}.

\begin{verbatim}
{
  "predicates": [
    {
      "ref": "p1",
      "kind": "condition",
      "payload": {
        "term": "type 2 diabetes",
        "count_min": 0
      },
      "polarity": "include"
    },
    {
      "ref": "p2",
      "kind": "condition",
      "payload": {
        "term": "hypertension",
        "count_min": 0
      },
      "polarity": "include"
    }
  ],
  "logic": {
    "ref": "g1",
    "operator": "OR",
    "children": [
      "p1",
      "p2"
    ]
  },
  "relations": []
}
\end{verbatim}

\begin{figure}[ht]
    \centering
    % \includegraphics[
    %     angle=90,
    %     origin=c,
    %     height=1.2\textwidth
    % ]{figures/initial.png}
    \makebox[\linewidth][c]{%
        \includegraphics[width=\linewidth]{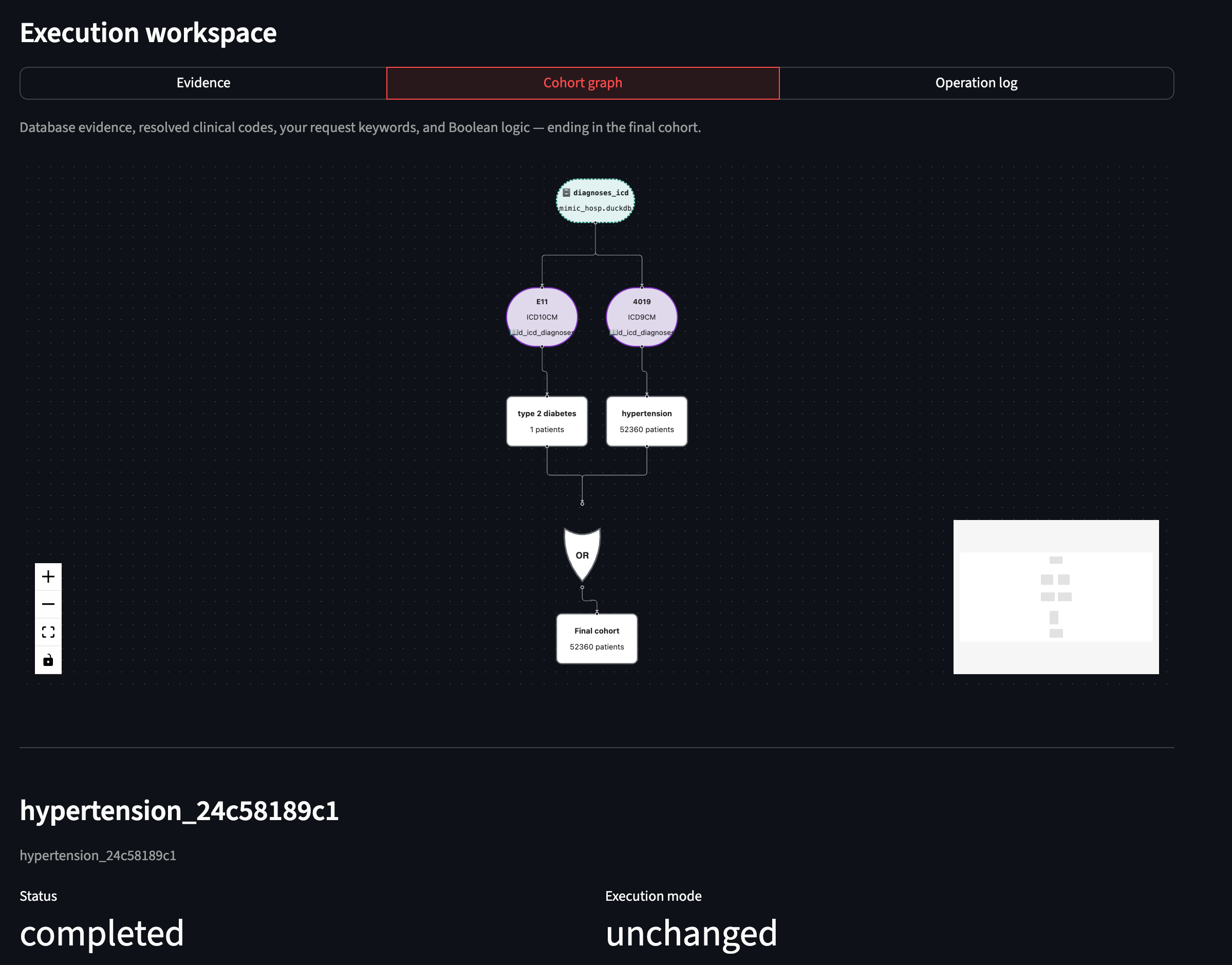}
    }
    \caption{Execution of the initial cohort request. The resolved type 2
    diabetes and hypertension branches are combined by the \textsc{or}
    operator to form the initial cohort.}
    \label{fig:appendix-initial}
\end{figure}

Figure~\ref{fig:appendix-initial} shows the corresponding execution after
clinical concept resolution. The two condition branches are evaluated
independently and combined according to the Boolean structure of the
specification.

\subsection{Localized Correction}

The user subsequently provides the correction:

\begin{quote}
\textit{``Patients must be under 40 years old.''}
\end{quote}

Rather than regenerating the complete specification, the correction interpreter
identifies a localized semantic change. For this correction, it produces the
following action:

\begin{verbatim}
{
  "actions": [
    {
      "action": "ADD",
      "local_ref": "new_1",
      "requirement_text":
        "patients must be under 40 years old",
      "excluded": false,
      "anchor": "ROOT",
      "operator": "AND",
      "relation_memberships": []
    }
  ],
  "requires_clarification": false,
  "clarification_question": null,
  "clarification_rationale": null
}
\end{verbatim}

The action introduces a new requirement at the root of the existing logical
structure. The requirement is compiled into predicate \texttt{p3} and attached
using conjunction, while the existing disjunction \texttt{g1} is retained.

\clearpage
\subsection{Revised Task Specification}

After deterministic reconciliation, the resulting \texttt{TaskSpec} is:

\begin{verbatim}
{
  "predicates": [
    {
      "ref": "p1",
      "kind": "condition",
      "payload": {
        "term": "type 2 diabetes",
        "count_min": 0
      },
      "polarity": "include"
    },
    {
      "ref": "p2",
      "kind": "condition",
      "payload": {
        "term": "hypertension",
        "count_min": 0
      },
      "polarity": "include"
    },
    {
      "ref": "p3",
      "kind": "patient",
      "payload": {
        "attribute": "age",
        "operator": "<",
        "value": 40
      },
      "polarity": "include"
    }
  ],
  "logic": {
    "ref": "g2",
    "operator": "AND",
    "children": [
      {
        "ref": "g1",
        "operator": "OR",
        "children": [
          "p1",
          "p2"
        ]
      },
      "p3"
    ]
  },
  "relations": []
}
\end{verbatim}

The revised logical expression is therefore
\[
    (p_1 \lor p_2) \land p_3.
\]
The original predicates and their disjunction remain structurally unchanged,
while the newly introduced age predicate requires new execution work.

\begin{figure}[ht]
    \centering
    \includegraphics[width=\linewidth]{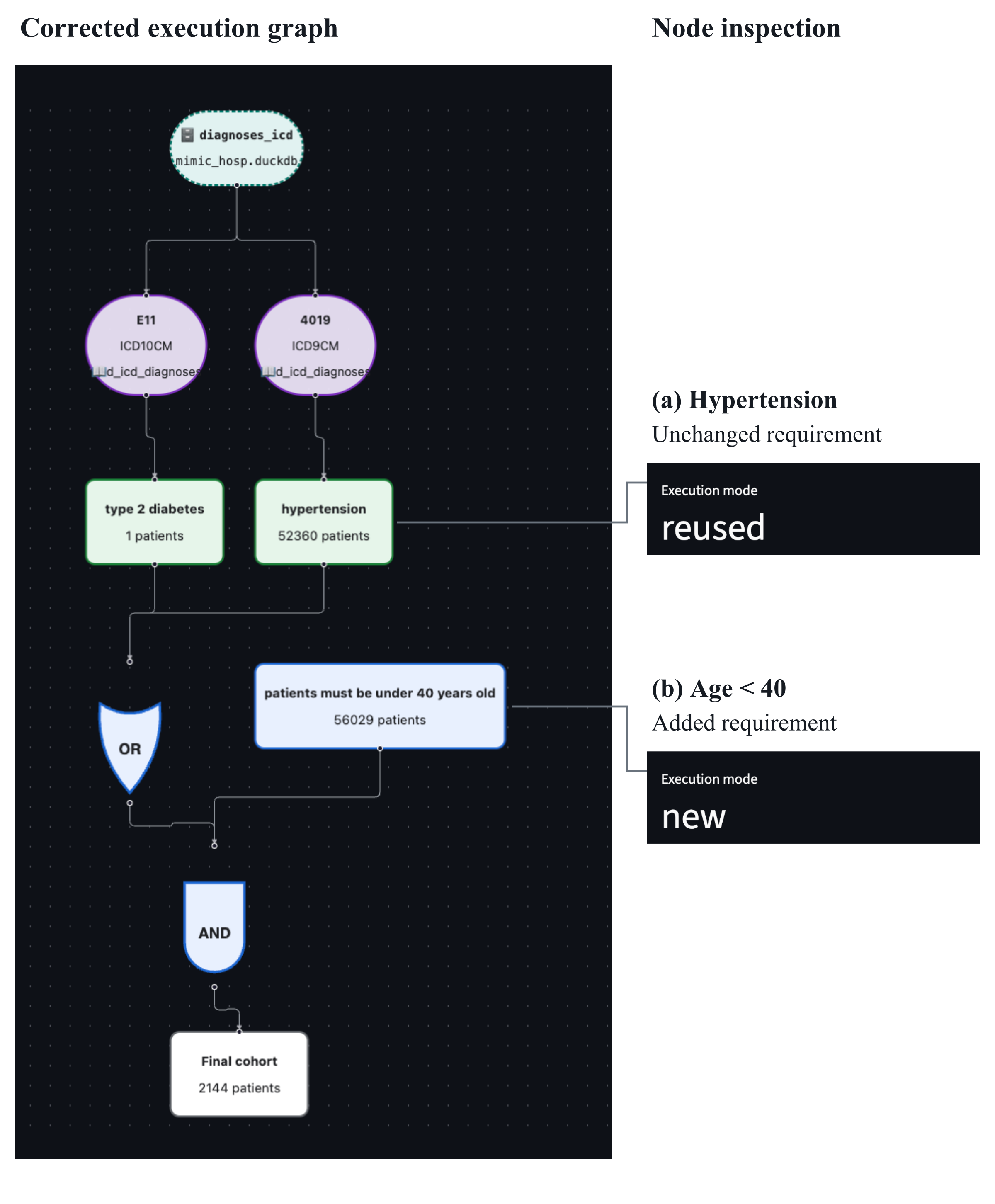}
    \caption{Execution after adding the constraint $\mathrm{age}<40$.
    The corrected graph preserves the original disjunction and combines it
    with the age predicate through \textsc{and}.
    Node inspection shows (a) reuse of the unchanged hypertension result
    and (b) execution of the age predicate as new work.}
    \label{fig:appendix-correction}
\end{figure}

Figure~\ref{fig:appendix-correction} shows the corrected execution graph
and node-level inspection of reused and newly executed work.

